\documentclass[acmconf]{acmart}

\AtBeginDocument{%
  \providecommand\BibTeX{{%
    \normalfont B\kern-0.5em{\scshape i\kern-0.25em b}\kern-0.8em\TeX}}}

\setcopyright{acmcopyright}
\copyrightyear{2026}
\acmYear{2026}
\acmDOI{}

\acmConference[ICICT 2026]{2026 The 9th International Conference on Information and Computer Technologies}{March 11--13, 2026}{Honolulu, HI}
\acmPrice{}
\acmISBN{}

\begin{document}

\title{RCDN: Real-Centered Detection Network for Robust Face Forgery Identification}

\author{Wyatt McCurdy}
\email{wyatt.mccurdy@maine.edu}
\affiliation{%
  \institution{University of Southern Maine}
  \streetaddress{96 Falmouth St.}
  \city{Portland}
  \state{Maine}
  \postcode{04103}
}

\author{Xin Zhang}
\affiliation{%
  \institution{University of Southern Maine}
  \streetaddress{96 Falmouth St.}
  \city{Portland}
  \state{Maine}
  \postcode{04103}
}
\email{xin.zhang@maine.edu}

\author{Yuqi Song}
\affiliation{%
  \institution{University of Southern Maine}
  \streetaddress{96 Falmouth St.}
  \city{Portland}
  \state{Maine}
  \postcode{04103}
}
\email{yuqi.song@maine.edu}

\author{Min Gao}
\email{gaomin@cqu.edu.cn}
\affiliation{%
  \institution{Chongqing University}
  \city{Chongqing}
}


\begin{abstract}
Image forgery has become a critical threat with the rapid proliferation of AI-based generation tools, which make it increasingly easy to synthesize realistic but fraudulent facial content. Existing detection methods achieve near-perfect performance when training and testing are conducted within the same domain, yet their effectiveness deteriorates substantially in cross-domain scenarios. This limitation is problematic, as new forgery techniques continuously emerge and detectors must remain reliable against unseen manipulations. To address this challenge, we propose the Real-Centered Detection Network (RCDN), a frequency–spatial convolutional neural networks(CNN) framework with an Xception backbone that anchors its representation space around authentic facial images. Instead of modeling the diverse and evolving patterns of forgeries, RCDN emphasizes the consistency of real images, leveraging a dual-branch architecture and a real-centered loss design to enhance robustness under distribution shifts. Extensive experiments on the DiFF dataset, focusing on three representative forgery types (FE, I2I, T2I), demonstrate that RCDN achieves both state-of-the-art in-domain accuracy and significantly stronger cross-domain generalization. Notably, RCDN reduces the generalization gap compared to leading baselines and achieves the highest cross/in-domain stability ratio, highlighting its potential as a practical solution for defending against evolving and unseen image forgery techniques.
\end{abstract}



% %% A "teaser" image appears between the author and affiliation
% %% information and the body of the document, and typically spans the

\maketitle

\section{Introduction}
The rapid advancement of generative artificial intelligence has elevated image forgery detection into a critical research area. With the emergence of powerful diffusion and generative adversarial models, forged images can now be produced with unprecedented realism and at minimal cost~\cite{croitoru2023diffusion}. These synthetic faces are not merely academic curiosities: they pose tangible threats to privacy, security, and trust in digital media. Malicious actors can exploit forged images for disinformation campaigns, identity theft, political manipulation, or social engineering attacks, undermining public confidence in visual evidence. Traditional forensic techniques, often relying on handcrafted features or shallow classifiers, are increasingly ineffective against such sophisticated fabrications, underscoring the urgent need for robust, generalizable detection methods.

\begin{figure}[!htbp]
    \centering
    \includegraphics[width=0.42\textwidth]{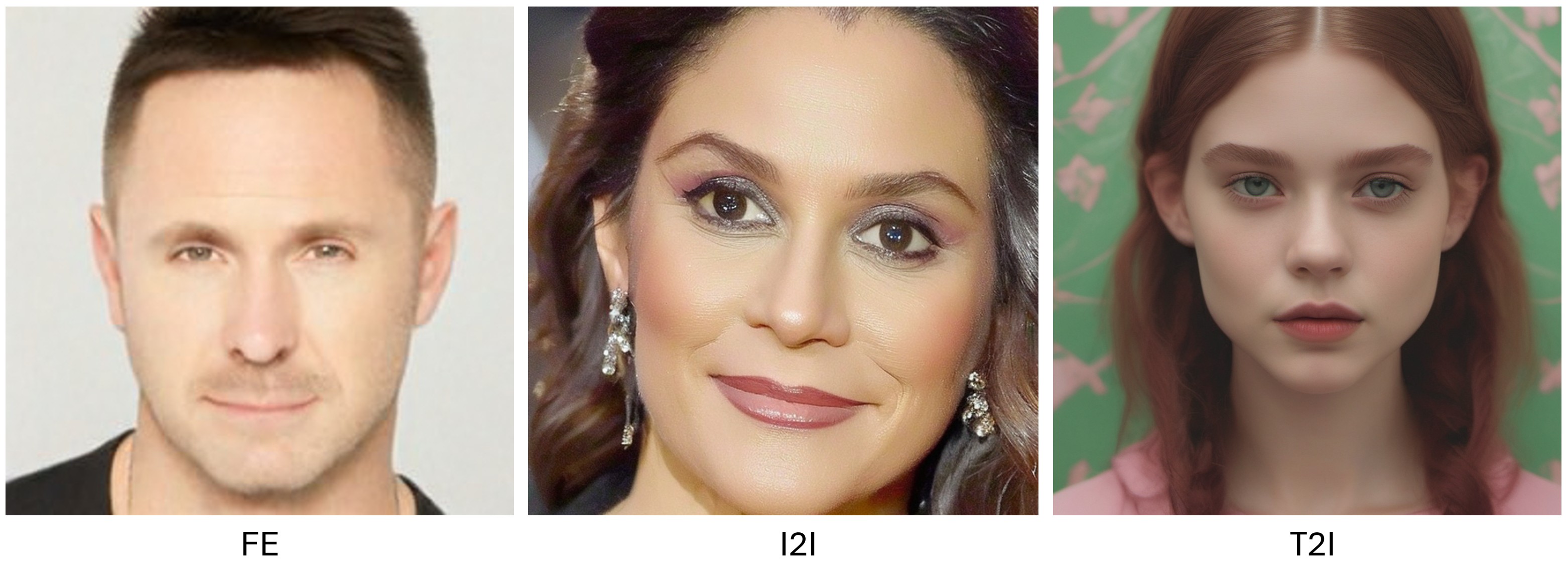}
    \Description{This figure shows three diffusion-generated images of a man's face on the left, a woman's face in the center and a woman's face on the right. The left face was generated by FE, the middle face was generated by I2I, and the right face was generated by T2I.}
    \caption{Examples of diffusion-based facial forgery categories considered in this work: Face Editing (FE) alters specific attributes of a real face, Image-to-Image translation (I2I) refines or transforms facial appearance from an input image, and Text-to-Image generation (T2I) creates a new face guided purely by textual descriptions.}
    \label{fig:repre}
\end{figure}

Recent research categorizes diffusion-based facial forgeries into several representative classes, among which Face Editing (FE)~\cite{shen2020interpreting}, Image-to-Image translation (I2I)~\cite{isola2017image}, and Text-to-Image generation (T2I)~\cite{ramesh2021zero} are the most practically relevant. FE manipulates specific attributes of an existing real face, such as altering age, expression, or hairstyle, while preserving the underlying identity. I2I uses a source face image as input and refines or transforms it into a new output that maintains overall resemblance but introduces subtle stylistic or identity-related changes. T2I, in contrast, does not require a source face at all: it synthesizes entirely new faces from textual prompts, enabling the creation of photorealistic but non-existent individuals. Together, these categories cover a broad spectrum of forgery mechanisms, from modifying authentic content to generating faces from scratch, and therefore capture the dominant ways diffusion models are used for face manipulation. Fig.~\ref{fig:repre} illustrates representative examples of these three forgery types.

Existing CNN-based detectors achieve strong in-domain performance, i.e., when trained and tested on forgeries of the same type~\cite{rossler2019faceforensics++,wang2020cnn}. However, their effectiveness deteriorates sharply in cross-domain scenarios, where the model is exposed to novel forgery methods or unseen distributions. This limitation is especially concerning: as forgery techniques evolve rapidly, detectors overfitted to a specific manipulation type fail to generalize, leaving systems vulnerable to new attacks~\cite{haliassos2021lips,chen2025prest}. Thus, improving cross-domain robustness is both urgent and necessary~\cite{dzanic2020fourier}.

To address this challenge, we propose the Real-Centered Detection Network (RCDN), a frequency–spatial CNN framework with an Xception backbone~\cite{chollet2017xception}. Instead of attempting to track the rapidly evolving patterns of fake images, RCDN emphasizes learning a compact and stable representation of real images. The key intuition is that, although the appearance of forgeries can differ drastically across generation methods, the distribution of authentic facial images remains consistent. Building on this insight, RCDN integrates a spatial branch that captures high-level semantic facial features with a frequency branch that models subtle spectral artifacts introduced by generative processes. A learnable real-center embedding further anchors the representation space, and the loss design enforces clustering of real samples around this center while driving forgeries away. By reinforcing the invariance of real images rather than chasing the variability of fakes, RCDN provides improved robustness against unseen forgery domains.

Our contributions can be summarized as follows:
\begin{itemize}
\item \textbf{Cross-domain detection strategy:} We propose RCDN, a real-centered CNN that combines spatial and frequency features to achieve generalization beyond in-domain forgery detection.
\item \textbf{Benchmarking on DiFF:} We evaluate RCDN on the DiFF benchmark, focusing on three representative forgery types (FE, I2I, T2I), and demonstrate consistent improvements in cross-domain performance.
\item \textbf{Open-source release:} To promote reproducibility and facilitate future research, we publicly release our implementation\footnote{https://github.com/wyattmccurdy12/Face-Forgery-Detection/tree/master}.
\end{itemize}

\section{Related Work}
\label{rw}
Image forgery detection has evolved from traditional signal processing approaches to sophisticated deep learning–based systems~\cite{farid2009image}. Early studies employed handcrafted features to address manipulations such as splicing, copy–move, and inpainting, making use of inconsistencies in noise, edges, or compression artifacts. These methods, while effective for simple manipulations, struggled as generative models advanced. With the emergence of deepfakes~\cite{mirsky2021creation} and diffusion-based synthesis~\cite{dabral2023mofusion}, CNN~\cite{o2015introduction} have become the predominant tools. Representative architectures such as Xception~\cite{chollet2017xception}, F3-Net~\cite{sun2023f3}, EfficientNet~\cite{koonce2021efficientnet}, and DIRE~\cite{wang2023dire} have been extensively applied to facial forgery detection and now serve as standard baselines. Beyond backbone CNNs, more recent research explores frequency-domain analysis to capture generative artifacts, attention mechanisms to emphasize discriminative regions, and multi-branch frameworks that fuse spatial and frequency information~\cite{luo2021generalizing}. Together, these approaches have significantly advanced the detection of forged content and established a strong methodological foundation for modern visual forensics.

\subsubsection{Diffusion-based facial forgery benchmarks} 
Many existing datasets for diffusion-generated images focus on generic scenes such as bedrooms and kitchens, while facial forgeries, which pose greater security and privacy risks, remain insufficiently represented~\cite{chen2024drct}. Although several facial datasets have been introduced, they are often small in scale, sometimes containing only about 1,500 images, and are collected under restricted conditions with limited prompts, which reduces their diversity and applicability. To overcome these limitations, the DiFF dataset~\cite{cheng2024diffusion} was developed as the first large-scale benchmark dedicated to diffusion-based facial forgery detection. It contains more than 500,000 forged facial images generated by thirteen state-of-the-art diffusion models, using over 20,000 textual prompts and 10,000 visual prompts spanning more than 1,000 identities. Each image is carefully annotated with its generation method and corresponding prompts, enabling detailed evaluation across multiple manipulation categories. DiFF therefore provides both a comprehensive resource for diffusion-generated facial forgery research and a robust benchmark for advancing detection methods.

\subsubsection{Cross-domain generalization}
While CNN-based detectors achieve strong in-domain accuracy, their performance declines substantially in cross-domain evaluations, where training and testing manipulations differ. Fine-tuning and linear probing partially mitigate this gap, and Edge Graph Regularization (EGR) has recently been proposed as a means of improving robustness~\cite{cheng2024diffusion}. However, these improvements remain modest, and cross-domain generalization continues to represent a central open challenge~\cite{haliassos2021lips,chen2025prest}. 
In response to this challenge, our study focuses on enhancing cross-domain performance in facial forgery detection. By achieving stronger generalization across manipulation categories, our approach demonstrates the potential to resist unseen or newly emerging forgery techniques, which is essential given the rapid evolution of generative models.

\section{Proposed Methods}
\label{pm}
Traditional forgery detection models directly learn to classify fake images. This works well in-domain but fails cross-domain, because fake distributions shift dramatically across forgery methods. In contrast, the distribution of real images is stable and consistent across domains. Based on this insight, we design RCDN to anchor its representation space around real images, so that the model generalizes better to unseen forgery methods.

\begin{figure*}[htbp]
    \centering
    \Description{This image shows a graphical representation of the network architecture of RCDN. The input is fed to a frequency branch on top and a spatial branch on bottom. There is then a graphic of the feature vectors being concatenated and subjected to our three loss functions, with calculation of distance from the real-center embedding shown.}
    \includegraphics[width=0.9\textwidth]{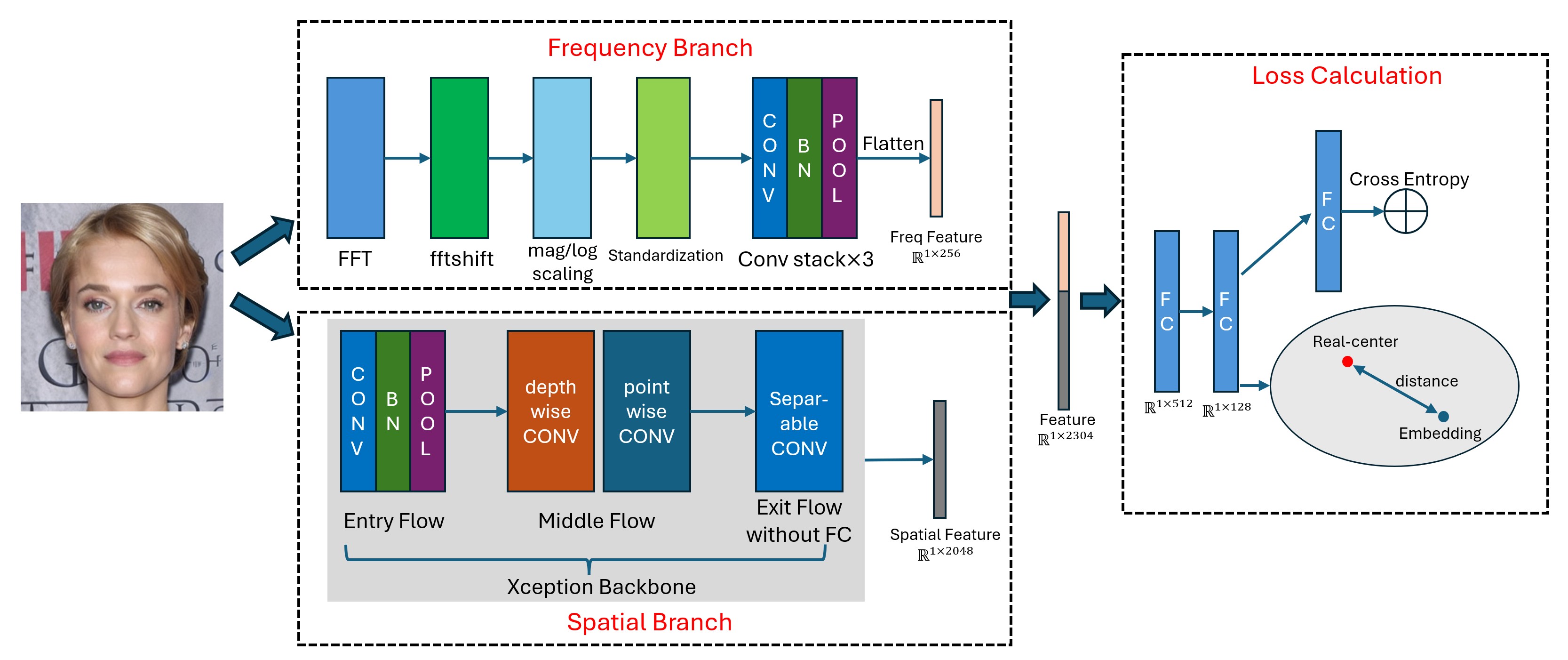}
    \caption{Overview of the proposed Real-Centered Detection Network (RCDN). The model processes an input face image through a frequency branch (FFT-based ConvNet) and a spatial branch (Xception backbone), concatenates the resulting features, and projects them into a 128-D embedding space. This embedding is supervised both by a classifier and by real-centered constraints. By anchoring the feature space around real images, RCDN achieves robustness across forgery domains.}
    \label{fig:pipe}
\end{figure*}

As shown in Fig.~\ref{fig:pipe}, the input face image is processed by two complementary branches. The frequency branch captures subtle spectral artifacts introduced during image generation via Fast Fourier
Transform (FFT)-based preprocessing followed by a lightweight ConvNet~\cite{liu2022convnet}, producing a 256-D feature. The spatial branch leverages an Xception backbone (without the classifier layer) to extract semantic facial features, resulting in a 2048-D feature. These are concatenated into a 2304-D vector and projected by a multilayer perceptron into a compact 128-D embedding, normalized for stability.
This embedding is used in two ways: (1) a fully connected classifier optimized with cross-entropy to distinguish real vs. fake; and (2) distance-based constraints with a learnable real-center embedding. The center loss pulls real embeddings closer to the center while pushing fake embeddings away, and the separation loss further enlarges the margin between the two distributions. Together, these losses enforce consistency around real images rather than variability of forgeries, addressing the cross-domain generalization challenge.

\subsection{Model Architecture}
The proposed \textbf{Real-Centered Detection Network (RCDN)} is designed to learn robust and transferable representations by jointly modeling spatial and frequency information, while explicitly anchoring its embedding space around authentic images. The architecture consists of two parallel feature extraction branches, a projection module, and dual supervision heads. The design reflects the motivation that although forgery methods evolve rapidly and exhibit diverse artifacts, the distribution of real images remains consistent across domains. Anchoring the feature space around real images therefore provides a principled pathway toward improved cross-domain robustness.

\subsubsection{Spatial branch}
The spatial branch is responsible for extracting high-level semantic representations of the input face image $x \in \mathbb{R}^{H \times W \times 3}$. We employ an Xception backbone, which is a deep convolutional architecture built upon depthwise separable convolutions. This design decomposes the conventional convolution into two steps: a depthwise operation applied channel-wise, followed by a pointwise convolution for feature mixing. Such factorization reduces computational cost while preserving representational power, making Xception well suited for large-scale feature extraction.  

The architecture of Xception can be divided into three stages. The \emph{entry flow} consists of standard and separable convolutional layers that reduce spatial resolution while capturing low-level textures and edges. The \emph{middle flow} is composed of multiple repeated blocks of depthwise separable convolutions, which form the core of the network and extract mid- to high-level semantic cues such as facial parts and structural patterns. Finally, the \emph{exit flow} applies separable convolutions followed by global average pooling to summarize information into a compact feature descriptor.  

Formally, we denote the spatial feature extractor as $\phi_{s}(\cdot)$, yielding a pooled feature vector:
\begin{equation}
f_{s} = \phi_{s}(x), \quad f_{s} \in \mathbb{R}^{1 \times d_s},
\end{equation}
where $d_s = 2048$. This vector encodes semantic and structural properties of the input face, serving as a robust foundation for distinguishing authentic from manipulated content.

\subsubsection{Frequency branch}
Generative manipulations often leave subtle traces in the frequency domain. To expose these cues, we transform the input image to the spectral domain using the two-dimensional Fourier transform, yielding complex coefficients $F(u,v)$ over horizontal and vertical frequencies. In the discrete Fourier transform, the zero-frequency component is typically located at the spectrum corners due to indexing conventions. For analysis and subsequent convolutional processing, it is advantageous to recenter the spectrum so that low frequencies lie at the geometric center and high frequencies radiate outward. We denote this index-recentering as a frequency shift operation,
\[
\widetilde{F}(u,v) = \mathcal{S}\big(F(u,v)\big),
\]
which is a permutation of indices that moves the zero-frequency component to the center of the array.

From the centered spectrum $\widetilde{F}(u,v)$, we compute the magnitude and apply logarithmic compression to balance the dominance of low frequencies and enhance mid-to-high frequency details:
\begin{equation}
\widetilde{M}(u,v) = \log\!\big(1 + \lvert \widetilde{F}(u,v) \rvert \big).
\end{equation}
Channel-wise standardization is then applied to normalize scale across RGB channels. The resulting spectral representation is processed by a compact convolutional subnetwork (several convolution, normalization, nonlinearity, and pooling layers), producing a low-dimensional vector that summarizes spectral regularities and anomalies:
\begin{equation}
f_{f} = \phi_{f}\!\big(\widetilde{M}\big), \qquad f_{f} \in \mathbb{R}^{1 \times d_f}, \; d_f = 256.
\end{equation}
Centering the spectrum concentrates energy relevant to global structure near the middle of the feature map and arranges higher frequencies with approximate radial symmetry, which improves locality for CNN filters and makes synthesis artifacts more accessible to the downstream convolutional extractor.

\subsubsection{Feature fusion and projection}
The outputs of the two branches are concatenated to form a joint representation:
\begin{equation}
f_{c} = [f_{s} \,\|\, f_{f}] \in \mathbb{R}^{1 \times (d_s + d_f)}.
\end{equation}
To obtain a compact and discriminative embedding, this vector is passed through a projection module consisting of fully connected layers with non-linear activation, yielding a 128-dimensional feature:
\begin{equation}
z = \psi(f_{c}), \quad z \in \mathbb{R}^{1 \times d_e}, \; d_e = 128.
\end{equation}
For training stability, the embedding is normalized onto a hypersphere:
\begin{equation}
\hat z = \frac{z}{\|z\|_2}.
\end{equation}

\subsubsection{Supervision heads}
The normalized embedding $\hat z$ constitutes the central representation of RCDN and is simultaneously supervised by two complementary heads:  

\begin{itemize}
    \item Classification head. A linear projection maps the embedding to logits for binary classification (real vs. fake). This component, trained with cross-entropy loss, ensures that the embedding captures discriminative cues within the training distribution.  

    \item Real-centered head. A learnable vector $c \in \mathbb{R}^{1 \times d_e}$ represents the prototype of authentic images. Real samples are constrained to lie close to this center, while fake samples are encouraged to be pushed away, regardless of their generative origin. By explicitly anchoring the feature space around real images, this mechanism enforces geometric consistency and enhances cross-domain generalization.  
\end{itemize}

\subsection{Loss Function Design}
RCDN is supervised by a combination of classification and geometry-based losses. This design ensures that the learned embedding is both discriminative within the training domain and robust to domain shifts. The final training objective can be expressed as a weighted sum:
\begin{equation}
\mathcal{L} = \mathcal{L}_{cls} + \lambda_{c} \mathcal{L}_{center} + \lambda_{s} \mathcal{L}_{sep},
\end{equation}
where $\mathcal{L}_{cls}$ denotes the cross-entropy classification loss, $\mathcal{L}_{center}$ is the real-centered loss, $\mathcal{L}_{sep}$ is the separation loss, and $\lambda_{c}, \lambda_{s}$ are balancing coefficients.

\subsubsection{Classification loss}
The classification head provides direct supervision by encouraging the embedding to distinguish between real and fake images. This component is trained with the standard cross-entropy loss, which penalizes incorrect predictions and ensures that the embedding remains discriminative within the training distribution.

\subsubsection{Real-centered loss}
To explicitly anchor the embedding space around real images, we introduce a learnable real-center vector $c \in \mathbb{R}^{1 \times d_e}$. For each normalized embedding $\hat{z}$, we compute its Euclidean distance to the center:
\begin{equation}
d(\hat{z}, c) = \lVert \hat{z} - c \rVert_2.
\end{equation}
The center loss enforces real samples to cluster tightly around the center while pushing fake samples away by at least a margin $m$:
\begin{equation}
\mathcal{L}_{center} = \frac{1}{|R|}\sum_{\hat{z}_r \in R} d(\hat{z}_r, c)^2 \;+\; 
\frac{1}{|F|}\sum_{\hat{z}_f \in F} \max\big(0, m - d(\hat{z}_f, c)\big),
\end{equation}
where $R$ and $F$ denote the sets of embeddings from real and fake images from one batch, respectively. This encourages the geometry of the embedding space to remain consistent across domains by modeling the stable distribution of real faces.

\subsubsection{Separation loss}
While the center loss constrains each sample relative to the center, the separation loss enforces a stronger condition at the level of distribution statistics. Let $\bar{d}_{real}$ and $\bar{d}_{fake}$ denote the mean distances of real and fake embeddings in the batch, respectively. The separation loss is defined as:
\begin{equation}
\mathcal{L}_{sep} = \max \big( 0, \, \bar{d}_{real} - \bar{d}_{fake} + m \big).
\end{equation}
This ensures that, on average, fake samples remain farther from the real center than real samples by at least a margin $m$. By explicitly separating batch-level expectations, this term strengthens global robustness against unseen forgery types.

These three components form a complementary training signal. The cross-entropy term encourages discriminability within the training distribution, while the real-centered and separation losses impose geometric structure that emphasizes the invariance of real images. By anchoring the embedding space around real faces and ensuring that forgeries are consistently displaced, RCDN is better equipped to resist domain shifts and generalize to novel manipulation techniques.

\section{Experiments}
\label{exp}

\subsection{Dataset}
The DiFF dataset provides a massive collection of diffusion-generated facial forgeries, with over 500,000 manipulated images spanning thirteen different generation models and a wide variety of prompts. While this scale is valuable, not all samples pose the same level of challenge for forgery detection. In practice, a substantial portion of generated images can be easily distinguished from real ones due to obvious artifacts, low-resolution details, or unrealistic facial features. Training on such samples risks inflating performance without truly testing generalization. To ensure a more meaningful and challenging evaluation, we focus on three representative categories, Face Editing (FE), Image-to-Image translation (I2I), and Text-to-Image generation (T2I), and deliberately subsample the data. Specifically, we select 10,000 training images and 2,000 testing images per category, emphasizing diverse but non-trivial cases. This subset not only reduces computational cost but also better reflects the core challenge of cross-domain detection, where subtle manipulations and realistic synthesis make fake images far harder to separate from authentic ones. The representative subset selection we adopt is consistent with the protocol reported in the official DiFF code repository, ensuring comparability with prior usage and reproducibility of our results. The details of our representative subset selection are documented in our accompanying code repository to ensure transparency and reproducibility.

\subsection{In-domain Performance}

Table \ref{tab:indomain} presents the in-domain detection accuracy, where models are trained and evaluated on disjoint splits originating from the same DiFF subset. 

\begin{table}[htbp]
\caption{In-domain detection accuracy on the DiFF dataset. Training and testing are conducted on the same subset.}
\centering
\begin{tabular}{|l|c|c|c|}
\hline
\textbf{Method} & \textbf{Train FE} & \textbf{Train I2I} & \textbf{Train T2I} \\
\hline
Xception~\cite{chollet2017xception}     & 0.9895 & 0.9860 & 0.9905 \\\hline
EfficientNet~\cite{koonce2021efficientnet} & 0.9980 & 0.9880 & 0.9930 \\\hline
ResNet+CBAM~\cite{maarefdoust2024attention}   & 0.9945 & 0.9775 & 0.9790 \\\hline
ResNet-34~\cite{xie2017aggregated}    & 0.9890 & 0.9835 & 0.9900 \\\hline
XcepKNN~\cite{gilles2025xcepknn}       & 0.8132 & 0.7773 & 0.7803 \\\hline
F3-Net~\cite{sun2023f3}        & 0.9940 & 0.9875 & 0.9930 \\\hline
DIRE~\cite{wang2023dire}          & 0.9820 & 0.9655 & 0.9850 \\\hline
\textbf{RCDN (Ours)} & 0.9995 & 0.9975 & 0.9990 \\
\hline
\end{tabular}
\label{tab:indomain}
\end{table}

In this setting, most contemporary CNN-based detectors achieve high accuracy, typically above 0.98, as the underlying forgery category remains consistent between training and testing. Xception, EfficientNet, ResNet variants, F3-Net, and DIRE all exhibit strong results, reflecting their capacity to capture category-specific manipulation artifacts when the generative process is known in advance.
Our proposed RCDN attains the highest accuracy across all three categories, consistently exceeding 0.997 and marginally outperforming state-of-the-art baselines. These findings indicate that, although in-domain evaluation is comparatively straightforward, RCDN is able to learn more discriminative representations of authentic versus manipulated images, thereby establishing a strong foundation for the more challenging cross-domain setting.

\subsection{Cross-domain Performance}

Table \ref{tab:crossdomain_raw} presents the cross-domain detection results, where training and testing are conducted on different manipulation categories.

Each row corresponds to a training subset (FE, I2I, or T2I), while each column shows the corresponding test subset. The diagonal entries, which represent in-domain evaluations, are omitted, and the rightmost column reports the average of the off-diagonal results for each training subset. This format allows a direct view of how well a detector trained on one forgery type generalizes to other unseen types.

\begin{table}[htbp]
\caption{Cross-domain detection accuracy on the DiFF dataset. Rows correspond to training subsets, and columns to testing subsets. In-domain diagonals are omitted; the rightmost column reports the cross-domain average.}
\centering
\begin{tabular}{|l|c|c|c|c|}
\hline
\textbf{Method (Train)} & \textbf{Test FE} & \textbf{Test I2I} & \textbf{Test T2I} & \textbf{Cross Avg} \\
\hline
Xception (FE)   & ---    & 0.8950 & 0.9115 & 0.9033 \\
Xception (I2I)  & 0.8380 & ---    & 0.9770 & 0.9075 \\
Xception (T2I)  & 0.8010 & 0.9590 & ---    & 0.8800 \\
\hline
EfficientNet (FE)   & ---    & 0.8960 & 0.9465 & 0.9213 \\
EfficientNet (I2I)  & 0.8605 & ---    & 0.9875 & 0.9240 \\
EfficientNet (T2I)  & 0.7810 & 0.9635 & ---    & 0.8773 \\
\hline
ResNet+CBAM (FE)   & ---    & 0.9095 & 0.9190 & 0.9142 \\
ResNet+CBAM (I2I)  & 0.8835 & ---    & 0.9685 & 0.9260 \\
ResNet+CBAM (T2I)  & 0.7785 & 0.9500 & ---    & 0.8642 \\
\hline
ResNet-34 (FE)   & ---    & 0.8695 & 0.8875 & 0.8785 \\
ResNet-34 (I2I)  & 0.8415 & ---    & 0.9760 & 0.9088 \\
ResNet-34 (T2I)  & 0.8175 & 0.9620 & ---    & 0.8898 \\
\hline
XcepKNN (FE)   & ---    & 0.6480 & 0.6164 & 0.6322 \\
XcepKNN (I2I)  & 0.6224 & ---    & 0.7410 & 0.6817 \\
XcepKNN (T2I)  & 0.5625 & 0.7638 & ---    & 0.6632 \\
\hline
F3-Net (FE)   & ---    & 0.8930 & 0.9075 & 0.9003 \\
F3-Net (I2I)  & 0.8260 & ---    & 0.9725 & 0.8992 \\
F3-Net (T2I)  & 0.7925 & 0.9550 & ---    & 0.8738 \\
\hline
DIRE (FE)   & ---    & 0.8930 & 0.9075 & 0.9003 \\
DIRE (I2I)  & 0.8960 & ---    & 0.9675 & 0.9318 \\
DIRE (T2I)  & 0.8185 & 0.9460 & ---    & 0.8823 \\
\hline
\textbf{RCDN (FE)}   & ---    & 0.9005 & 0.9685 & 0.9345 \\
\textbf{RCDN (I2I)}  & 0.8975 & ---    & 0.9980 & 0.9478 \\
\textbf{RCDN (T2I)}  & 0.8595 & 0.9970 & ---    & 0.9283 \\
\hline
\end{tabular}
\label{tab:crossdomain_raw}
\end{table}

A clear trend emerges across existing baselines: although they achieve near-perfect accuracy in in-domain evaluation, their performance drops notably when applied to unseen forgery types. For instance, Xception falls from nearly 0.99 in-domain to an average of 0.897 across domains, and EfficientNet drops to 0.877–0.924 depending on the training subset. Similar degradations are observed in ResNet variants and other strong baselines, indicating that current detectors often overfit to forgery-specific artifacts and struggle to generalize when encountering new generative processes. By contrast, our proposed RCDN maintains consistently high accuracy, exceeding 0.90 in all cross-domain settings and reaching nearly 0.99 when transferring from I2I to T2I, demonstrating clear superiority in handling distribution shifts.

Table \ref{tab:generalization} provides a statistical summary of the in-domain and cross-domain results.

For each method, the in-domain accuracy is computed as the average of diagonal entries in Table \ref{tab:indomain}, where training and testing are conducted on the same subset. The cross-domain average is obtained by averaging the off-diagonal entries in Table \ref{tab:crossdomain_raw}, corresponding to evaluations where training and testing are performed on different forgery categories. Based on these two values, we report two indicators of generalization. The gap is defined as the absolute difference between in-domain and cross-domain averages, reflecting how much accuracy degrades when moving from familiar to unseen forgery types. The ratio is calculated as the cross-domain average divided by the in-domain average, serving as a normalized measure of stability across domains.

\begin{table}[htbp]
\caption{Generalization analysis of in-domain and cross-domain performance on the DiFF dataset. ``Cross Avg'' is the mean of off-diagonal results, ``Gap'' is the accuracy difference (In-domain $-$ Cross Avg), and ``Ratio'' is the relative generalization (Cross Avg / In-domain).}
\centering
\begin{tabular}{|l|c|c|c|c|}
\hline
\textbf{Method} & \textbf{In-domain} & \textbf{Cross Avg} & \textbf{Gap} & \textbf{Ratio} \\
\hline
Xception~\cite{chollet2017xception}       & 0.9887 & 0.8970 & 0.0917 & 0.907 \\\hline
EfficientNet~\cite{koonce2021efficientnet}  & 0.9930 & 0.9075 & 0.0855 & 0.914 \\\hline
ResNet+CBAM~\cite{maarefdoust2024attention}   & 0.9837 & 0.9015 & 0.0822 & 0.916 \\\hline
ResNet-34~\cite{xie2017aggregated}     & 0.9875 & 0.8924 & 0.0951 & 0.904 \\\hline
XcepKNN~\cite{gilles2025xcepknn}        & 0.7903 & 0.6590 & 0.1313 & 0.834 \\\hline
F3-Net~\cite{sun2023f3}         & 0.9915 & 0.8911 & 0.1004 & 0.898 \\\hline
DIRE~\cite{wang2023dire}           & 0.9775 & 0.9048 & 0.0727 & 0.926 \\\hline
\textbf{RCDN (Ours)} & \textbf{0.9987} & \textbf{0.9369} & \textbf{0.0618} & \textbf{0.938} \\
\hline
\end{tabular}
\label{tab:generalization}
\end{table}

This analysis highlights the limitations of existing approaches: although most methods achieve near-perfect accuracy in-domain, their cross-domain averages drop by 8–10 percentage points, yielding ratios around 0.90. By contrast, our proposed RCDN shows both the smallest gap (0.0618) and the highest ratio (0.938), demonstrating that its accuracy under distribution shifts remains closer to its in-domain baseline than any competing method. These results confirm that RCDN achieves not only strong absolute performance but also superior generalization stability, a crucial property for deployment against evolving and previously unseen forgery techniques.

In summary, the cross-domain results demonstrate three key points: (1) most existing detectors suffer significant performance degradation when tested on unseen forgery types, (2) specialized architectures such as DIRE provide moderate robustness but still exhibit notable gaps, and (3) RCDN achieves both the strongest average performance and the most stable generalization, making it well-suited for resisting evolving forgery techniques in practice.

\section{Conclusion}
\label{con}

In this work, we addressed the challenge of cross-domain robustness in image forgery detection, an increasingly important problem given the rapid evolution of generative technologies. While most existing detectors achieve near-perfect performance in in-domain settings, their accuracy degrades substantially when evaluated on unseen forgery types. To mitigate this limitation, we proposed the Real-Centered Detection Network (RCDN), which anchors its representation space around real images rather than directly modeling the variability of forgeries. By integrating a spatial branch for semantic cues, a frequency branch for spectral artifacts, and a real-centered embedding with tailored loss functions, RCDN achieves both state-of-the-art accuracy and superior stability across domains.
Extensive experiments on the DiFF benchmark demonstrate that RCDN consistently outperforms strong baselines. Notably, it exhibits the smallest generalization gap and the highest cross/in-domain ratio, confirming its resilience to distribution shifts. These results indicate that focusing on the consistency of real images provides a promising direction for combating future, unseen forgery techniques.

\bibliographystyle{ACM-Reference-Format}
\bibliography{ref}

\end{document}